%% file: main.tex
\documentclass[preprint,review,3p,12pt,times,sort&compress]{elsarticle}

\usepackage{epsfig}
\usepackage{graphicx}
\usepackage{amsmath}
\usepackage{amssymb}
\usepackage{subfiles}
\usepackage{multirow} 

\usepackage{booktabs}
\usepackage[ruled]{algorithm2e}
\usepackage{arydshln}
\usepackage{bbding}
\usepackage{float}
\usepackage{utfsym}
\usepackage{CJKutf8}

\usepackage{makecell}
\usepackage{xcolor}
\usepackage{color, colortbl}
\definecolor{citecolor}{HTML}{0071bc}
\definecolor{tabhighlight}{HTML}{e5e5e5}
\usepackage[breaklinks=true,bookmarks=false]{hyperref}

\definecolor{diff}{rgb}{1,1,1}
\usepackage{soul}
\usepackage{mdframed}
\mdfsetup{%
linecolor=diff,
backgroundcolor=diff,
}
\soulregister{\cite}7
\soulregister{\citep}7
\soulregister{\citet}7
\soulregister{\ref}7
\soulregister{\pageref}7

\journal{Pattern Recognition}

\begin{document}

\begin{frontmatter}

%% Title, authors and addresses

%% use the tnoteref command within \title for footnotes;
%% use the tnotetext command for theassociated footnote;
%% use the fnref command within \author or \address for footnotes;
%% use the fntext command for theassociated footnote;
%% use the corref command within \author for corresponding author footnotes;
%% use the cortext command for theassociated footnote;
%% use the ead command for the email address,
%% and the form \ead[url] for the home page:
%% \title{Title\tnoteref{label1}}
%% \tnotetext[label1]{}
%% \author{Name\corref{cor1}\fnref{label2}}
%% \ead{email address}
%% \ead[url]{home page}
%% \fntext[label2]{}
%% \cortext[cor1]{}
%% \affiliation{organization={},
%%             addressline={},
%%             city={},
%%             postcode={},
%%             state={},
%%             country={}}
%% \fntext[label3]{}

%\title{Improving CLIP for Better Explainability with Enhancement in Open-Vocabulary Tasks}
\title{A Closer Look at the Explainability of Contrastive Language-Image Pre-training}
%\title{Enhancing the Explainability of Contrastive Language-Image Pre-training for Reliable Class Attention Map}

\author[label1]{Yi Li}
\author[label1]{Hualiang Wang}
\author[label2]{Yiqun Duan}
\author[label3]{Jiheng Zhang}
\author[label1]{Xiaomeng Li\corref{ca}}
\cortext[ca]{Corresponding author (eexmli@ust.hk).}
\affiliation[label1]{organization={Department of Electronic and Computer Engineering, The Hong Kong University of Science and Technology},
            %addressline={}, 
            city={Hong Kong},
            %postcode={}, 
            %state={},
            country={China}}
\affiliation[label2]{organization={School of Computer Science, University of Technology Sydney},%Department and Organization
            addressline={Ultimo}, 
            city={NSW},
            postcode={2007}, 
            %state={},
            country={Australia}}
\affiliation[label3]{organization={Department of Industrial Engineering and Decision Analytics, The Hong Kong University of Science and Technology},
            %addressline={}, 
            city={Hong Kong},
            %postcode={}, 
            %state={},
            country={China}}

\begin{abstract}
%CLIP is a powerful vision model that has shown great benefits for various tasks. However, we have identified some issues with its explainability, which affects its credibility and hinders related tasks. Specifically, CLIP tends to focus more on background regions rather than foregrounds, which goes against human understanding. Additionally, there are noisy activations in irrelevant positions on the visualization results. To address these problems, we conducted thorough analyses and discovered the reasons behind them. Building on these insights, we propose CLIP Surgery, a method that allows surgery-like modifications to the inference architecture and features without any fine-tuning. This approach significantly improves the explainability of CLIP for both convolutional networks and vision transformers, surpassing existing methods by a wide margin. Furthermore, our method also achieves remarkable improvements in open-vocabulary tasks, including multi-label recognition tasks (+4.41\% mAP on NUS-Wide), semantic segmentation (+8.74\% mIoU on Cityscapes), interactive segmentation based on SAM and multimodal visualization.

 Contrastive language-image pre-training (CLIP) is a powerful vision-language model that has shown great benefits for various tasks. However, we have identified some issues with its explainability, which undermine its credibility and limit the capacity for related tasks. Specifically, we find that CLIP tends to focus on background regions rather than foregrounds, with noisy activations at irrelevant positions on the visualization results. These phenomena conflict with conventional explainability methods based on the class attention map (CAM), where the raw model can highlight the local foreground regions using global supervision without alignment. To address these problems, we take a closer look at its architecture and features. Based on thorough analyses, we find the raw self-attentions link to inconsistent semantic regions, resulting in the opposite visualization. Besides, the noisy activations are owing to redundant features among categories. Building on these insights, we propose the CLIP Surgery for reliable CAM, a method that allows surgery-like modifications to the inference architecture and features, without further fine-tuning as classical CAM methods. This approach significantly improves the explainability of CLIP, surpassing existing methods by large margins. Besides, it enables multimodal visualization and extends the capacity of raw CLIP on open-vocabulary tasks without extra alignment. The code is available at https://github.com/xmed-lab/CLIP\_Surgery.

\end{abstract}


\end{frontmatter}

%\maketitle

\input{1_intro}
\input{2_ralated}
\input{3_method}
\input{4_experiments}
\input{5_conclusion}

%% The Appendices part is started with the command \appendix;
%% appendix sections are then done as normal sections
%% \appendix

%% \section{}
%% \label{}

%% If you have bibdatabase file and want bibtex to generate the
%% bibitems, please use
%%
\bibliographystyle{elsarticle-num} 
\bibliography{main}
%% else use the following coding to input the bibitems directly in the
%% TeX file.

% \begin{thebibliography}{00}

% %% \bibitem{label}
% %% Text of bibliographic item

% \bibitem{}
% \end{thebibliography}

\end{document}

%% file: 1_intro.tex
\section{Introduction}

%% CAM 很有用，可以做很多CNN的应用。最近， CLIP 非常popular, 可以做很多事情。however, CAM's application in CLIP is less explored. This paper focuses on "no training" - meaningful mask, enable downstream tasks. 

%% Related work. A navive using in CLIP. Related work calssfieid to two categories: (1) CLIP need training ---. However, xxxx  (2) 

Providing explanations for neural networks can enhance their transparency and credibility, which has become a crucial consideration across various fields of application. Among various explainability schemes \cite{baur2024explainability}, the class attention map (CAM) series methods ~\cite{zhou2016learning,selvaraju2017grad} explain the model via locating discriminative regions, which are widely used in applications such as semantic segmentation~\cite{li2021pseudo,yu2023ex}, image retrieval \cite{yu2024gradient} and generation~\cite{bar2022text2live}, etc. % However, it is less explored for the CAM of the contrastive language-image pre-training (CLIP)~\cite{radford2021learning}, a large vision-language pre-training model for varied tasks \cite{liu2022open,esmaeilpour2021zero,ramesh2022hierarchical,rombach2022high}. This paper aims to generate reliable CAM for CLIP, without any fine-tuning. Importantly, high-quality CAM enhances downstream tasks and plays a crucial role in promoting transparency. Furthermore, it provides valuable insights for understanding and improving the model.
Recently, contrastive language-image pre-training (CLIP) \cite{radford2021learning} has gained significant popularity and has been widely adopted in various downstream tasks such as segmentation \cite{xu2021simple}, generation \cite{bar2022text2live,tan2023semantic}. Although some methods achieve reasonable visualizations through additional modules and alignments~\cite{li2022exploring,huang2023generic}, they require further tuning \cite{zheng2024exploring} and do not provide a feasible CAM to explain the raw model of CLIP. Therefore, there is a strong need to develop a CLIP-CAM model that can enhance the model's transparency and credibility, enabling its \emph{direct application in various downstream tasks without requiring any fine-tuning and alignment.}

 % many CAM-based explainability methods \cite{zhou2016learning,selvaraju2017grad,lapuschkin2019unmasking,chefer2021generic,chefer2021transformer,chen2022gscorecam} yield poor visualizations; see Figure \ref{fig_vis_comp}.

\begin{figure}[H]
\centering
 \includegraphics[width=1\textwidth]{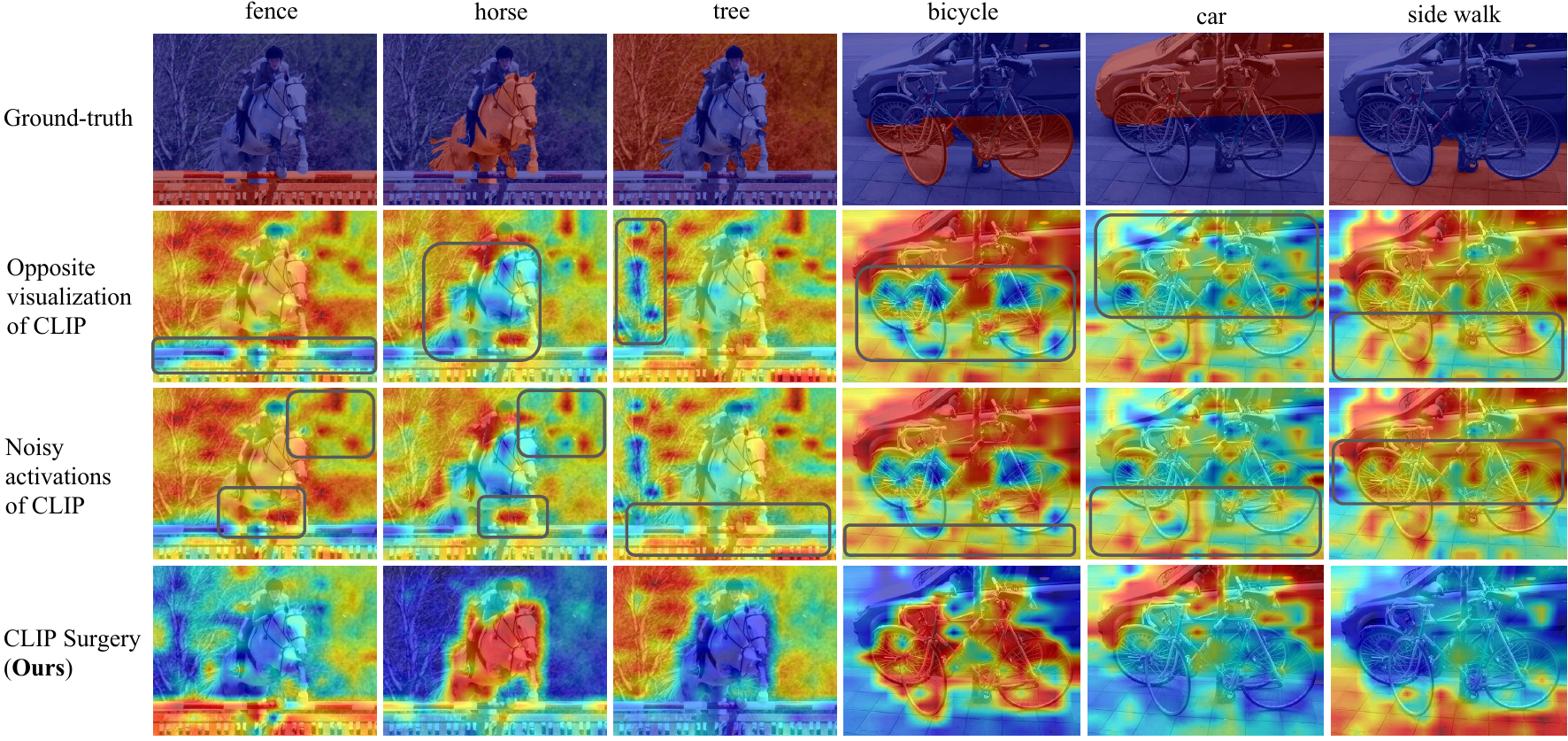}
\caption{The CAM of CLIP exhibits opposite and noisy visualizations, as indicated by the black boxes in the second and third rows, respectively. In contrast, our CLIP surgery addresses these issues, resulting in improved visualizations, as shown in the last row.
The foregrounds are shown in red, while the backgrounds are indicated in blue.}
\label{fig_problem}
\end{figure}

CAM explainability methods were initially designed for convolutional neural networks \cite{zhou2016learning,selvaraju2017grad} and vision transformers~\cite{chefer2021transformer}. However, when directly applied to CLIP, these methods yield unsatisfactory results as shown in Fig. \ref{fig_vis_comp}, including some recent works for multimodal models \cite{chefer2021generic} and CLIP object localization \cite{chen2022gscorecam}. This is mainly because the fundamental techniques do not work on CLIP without local alignment \cite{li2022exploring}, rendering subsequent improvement methods ineffective. Specifically, direct application of the basic CAM to CLIP reveals a tendency of CLIP to prioritize background regions over foregrounds, leading to "noisy activations" with "opposite visualizations" as shown in Figure \ref{fig_problem}. Besides, these phenomena also occur in the basic Grad-CAM \cite{selvaraju2017grad} as Fig. \ref{fig_vis_comp}, which is widely used in gradient-based CAM methods \cite{chefer2021generic,chen2022gscorecam}.

To address the observed explainability issues and generate high-quality CAM, we take a closer look at the architecture and features of CLIP to analyze how these phenomena happen. For the architecture, we observe that the self-attention layers build relations among inconsistent semantic regions (see. Fig. \ref{fig_token_shift}), resulting in the opposite visualization. Besides, not all the layers are beneficial and close to the final predictions as Fig. \ref{fig_cos_angle}. For the features, we find the noisy activations are usually irrelevant to labels, appearing with empty textual input (see Fig. \ref{fig_empty_label}), which suggests some features are redundant among categories. Based on these insights, we proposed the CLIP Surgery for reliable CAM, an approach that allows surgery-like modifications to the inference architecture and output features on the raw CLIP, called CLIP architecture surgery and CLIP feature surgery, respectively. To be specific, the architecture surgery is designed to solve the opposite visualization via reforming a consistent self-attention module using original parameters, and aggregate partial beneficial modules via a dual paths structure. Besides, we identify common features across classes as redundant features and mitigate them in the feature surgery to mitigate noisy activations.

\if 1
However, we notice that the raw predictions of CLIP behave differently from conventional CAM results, where CLIP pays more attention to background regions rather than foregrounds, with noisy activations at irrelevant positions, as shown in Fig. \ref{fig_problem}. Notably, many CAM-based explainability methods \cite{zhou2016learning,selvaraju2017grad,chefer2021generic，chen2022gscorecam} yield poor visualizations on CLIP, including methods based on gradient~\cite{selvaraju2017grad}, layerwise relevance propagation \cite{lapuschkin2019unmasking}, Bi-Modal~\cite{chefer2021generic} built upon explainability for vision transformer \cite{chefer2021transformer}, and gScoreCAM~\cite{chen2022gscorecam} for CLIP-based localization, as shown in Fig.~\ref{fig_vis_comp}. Besides these CAM-based methods, some recent works can achieve reasonable visualizations through additional fine-tuning and alignments, such as mask from DINO \cite{li2022exploring}, VQA-based alignment \cite{huang2023generic}, extra bounding box \cite{paiss2022no}. Nevertheless, these methods do not explain the original CLIP model. This is because they rely on additional processes such as fine-tuning, extra models, or layers, which can be less convenient, versatile, and practical compared to CAM methods that do not require fine-tuning.
\fi

Extensive experiments demonstrate the outstanding effectiveness of the proposed CLIP Surgery. Compared with CLIP in terms of explainability, the average mIoU improvements for varied backbones range from 22.11\% to 35.95\% on multiple datasets as Tab. \ref{tab_comp_clip}. Notably, the metric mSC (score difference between foregrounds and backgrounds) indicates CLIP prefers background than foreground, while our method solves this
issue with average mSC improvements over 47.72\%. It also significantly surpasses state-of-the-art CAM methods \cite{lapuschkin2019unmasking,chefer2021generic,chen2022gscorecam} as Tab. \ref{tab_comp_sota}) (e.g., more than 20\% mIoU improvements on VCO 2012 dataset \cite{everingham2010pascal}, even beyond that using extra alignment \cite{li2022exploring}. Besides, our method shows wide applicability on open-vocabulary tasks, such as semantic segmentation, interactive segmentation, multi-label recognition and multimodal visualization.

In summary, this paper has three main contributions: 

\begin{itemize}
    
    \item We observe that CLIP exhibits opposite visualization and noisy activations. Then, we discover that these phenomena are accompanied by inconsistent self-attention and redundant features among categories, respectively.
    
    \item Based on these insights, we propose the CLIP Surgery for reliable CAM, consisting of architecture and feature surgery without fine-tuning.

    \item The proposed method greatly improves the explainability of CLIP across various backbones and datasets, with wide applicability on multimodal visualization and open-vocabulary tasks.

\end{itemize}

%% file: 2_ralated.tex
\vspace{-0.1cm}
\section{Related Works}

\subsection{Explainability of CLIP}
Recently, CLIP has emerged as a powerful pre-training model supervised by natural language \cite{radford2021learning}. Before CLIP, traditional explainability methods such as CAM \cite{zhou2016learning}, Grad-CAM \cite{selvaraju2017grad}, etc. \cite{li2024cr} are designed for convolutional neural networks (CNNs). Recent methods~\cite{chefer2021transformer} have focused on explainability for vision transformers~\cite{dosovitskiy2020image} based on gradient. Besides, Bi-Modal \cite{chefer2021generic} and gScoreCAM \cite{chen2022gscorecam} were introduced for explainability on multimodal models and CLIP's object localization task, respectively. However, these class attention map (CAM) based methods show unsatisfactory results on CLIP as shown in Table~\ref{tab_comp_sota}. Besides the above CAM-based methods, some recent works generate reasonable visualizations via extra alignments, such as self-supervised mask \cite{li2022exploring}, VQA-based alignment \cite{huang2023generic}, extra bounding box \cite{paiss2022no} or prompt learning \cite{zheng2024exploring}. However, these methods do not explain the original CLIP model since the usage of extra fine-tuning, models, or
layers, and they are not as convenient and practical as our CAM-based methods. Besides, the proposed CLIP Surgery is much more effective on CLIP compared with existing CAM-based methods including CAM \cite{zhou2016learning} Grad-CAM \cite{selvaraju2017grad}, LRP \cite{lapuschkin2019unmasking}, Bi-Modal \cite{chefer2021generic}, gScoreCAM \cite{chen2022gscorecam} in Tab. \ref{tab_comp_sota}, with extensive abilities on downstream tasks. Besides, it surpasses alignment methods ECLIP \cite{li2022exploring}, etc. \cite{huang2023generic,paiss2022no} on practicalness and flexibility. Importantly, we aim to explain the explainability behaviors of the original CLIP model, while additional models, layers or fine-tuning in alignment methods violate this goal.

\subsection{Applications of CLIP} \label{sec2.2}
In general, CLIP is used as the pre-training model for downstream tasks such as zero-shot recognition~\cite{radford2021learning}, segmentation~\cite{xu2021simple}, detection \cite{huang2024joint}, generation~\cite{bar2022text2live,tan2023semantic} etc. These applications are achieved by task-specific designs out of the raw CLIP model. For open-vocabulary semantic segmentation, additional models or supervisions besides texts are used, such as extra fully-supervised proposal models \cite{zhang2023simple}. Besides, additional fine-tuning or training are involved, like token grouping in GroupViT \cite{xu2022groupvit} or self-training in MaskCLIP+ \cite{zhou2022extract}, and other sophisticated methods \cite{xu2023learning}. For interactive segmentation like segment anything model (SAM) \cite{kirillov2023segment}, it performs poorly with text prompts from CLIP alone, and the authors suggest combining text with manual points for better results. For CLIP based multi-label recognition, the alignment between visual and textual features are necessary for studies like Dual Modality \cite{xu2022dual}, DualCoOp \cite{sun2022dualcoop}, TaI-DPT \cite{guo2023texts}. 

Unlike these methods built on task-specific alignments, we observe that the raw CLIP without further alignment can achieve comparable performances on some applications with the help of our CLIP Surgery. Specifically, the high-quality CAM from our method can be directly used to generate segmentation results, or generating points from CAM to replace manual points for text-based SAM \cite{kirillov2023segment}. Besides, the performances of multi-label recognition are improved when we mitigate the redundant features in the feature surgery, even CAM-based methods \cite{zhou2016learning,selvaraju2017grad} are not responsible for classification improvements. Our method is also capable of multimodal visualizations to explain the learning process of CLIP. Notably, our method enhances the transparency of CLIP and enables it to serve multiple tasks simultaneously as shown in Fig. \ref{applicability}, beyond those fine-tuning methods for a certain specific task.

%% file: 3_method.tex
\section{Method} \label{sec3}
\begin{figure}[ht]
\centering
 \includegraphics[width=1\textwidth]{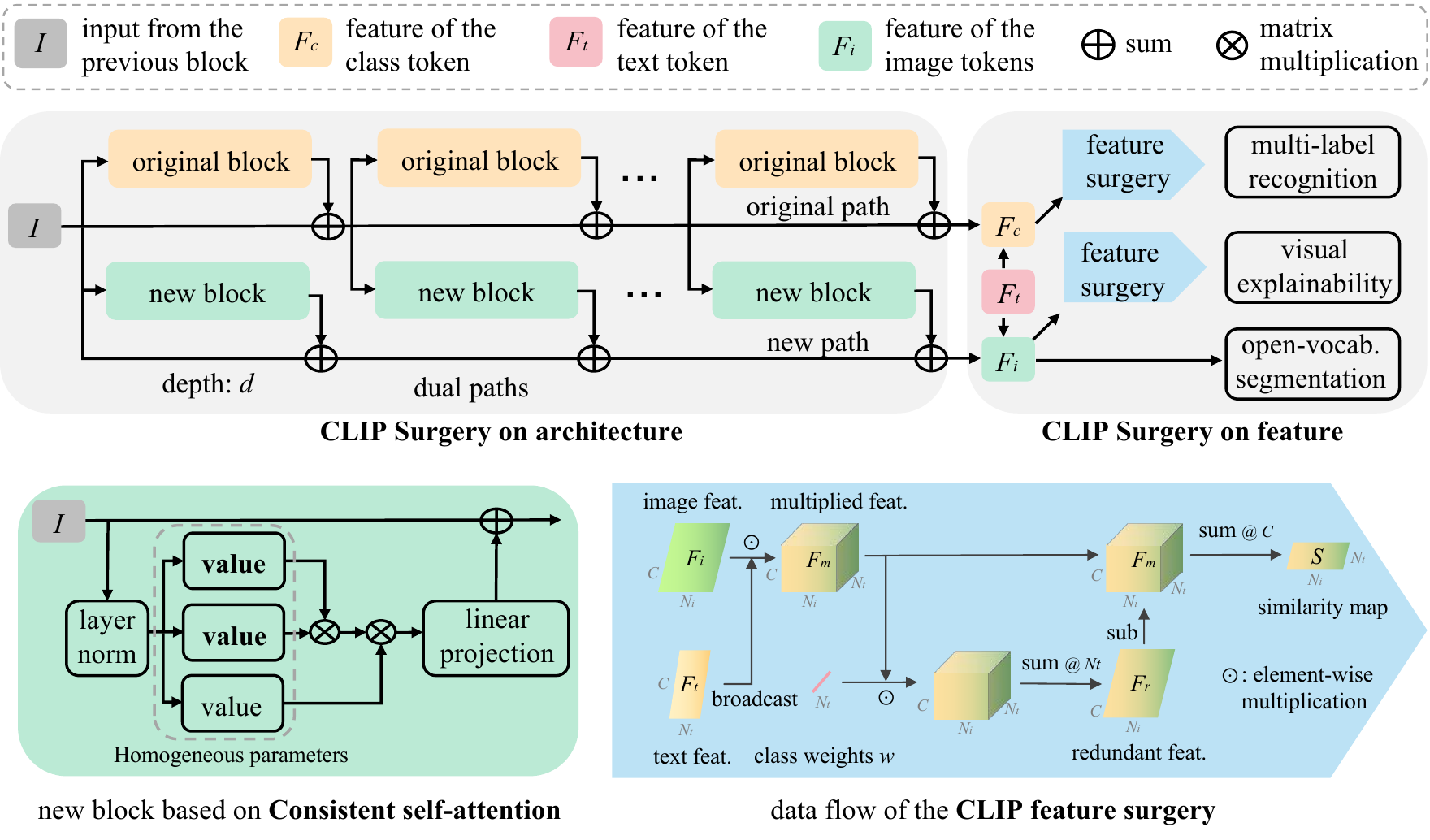}
\caption{The proposed CLIP Surgery contains two parts as the top part. For the architecture surgery, it is built on a dual paths structure, where the new blocks (left bottom part) are based on the consistent self-attention using homogeneous parameters without feed-forward networks (FFN). The data flow of feature surgery is given in the right bottom part to mitigate redundant features across texts.}
\label{fig_archi_surgery}
\end{figure}

In this section, we first introduce the visualization of CLIP from its raw predictions in Sec. \ref{sec3.1}, with descriptions of observed opposite visualization and noisy activation. Then, we analyze the architecture of CLIP in Sec. \ref{sec3.2}, and discover how the opposite visualization happens, with our solution: consistent self-attention and dual paths. In Sec. \ref{sec3.3}, we observe that the noisy activation is related to redundant features among categories, then propose the feature surgery to mitigate it. The overall framework of the CLIP Surgery is depicted in Fig. \ref{fig_archi_surgery} before detailed descriptions.

\subsection{Visual Explainability of CLIP} \label{sec3.1}

Foremost, we visually explain the CLIP from the similarity map as the class attention map from its raw predictions. Specifically, the raw predictions are the similarity distances between the text feature and image features of multiple image tokens. This similarity map is the most fundamental and direct explainability cue of CLIP, since it does not require any extra operation like back-propagation in previous gradient-based CAM methods \cite{selvaraju2017grad,chefer2021transformer}. Here we define the similarity map $\boldsymbol{M} \in \mathbb{R}^{H \times W \times N_t}$ as:
\begin{equation}
\label{eq_sm}
\boldsymbol{M} = \mathcal{N}(\mathcal{I}(\mathcal{R}(\boldsymbol{F}_{i} \boldsymbol{F}_{t}^{\top}))),
\end{equation}
where the L2-normalized image feature $\boldsymbol{F}_{i} \in \mathbb{R}^{N_i \times C}$ and transposed text feature $\boldsymbol{F}_{t}^{\top} \in \mathbb{R}^{N_t \times C}$ are multiplied to get the similarity map ($N_i$, $N_t$, $C$ are number of image token, text token and channel, respectively). Note this similarity map is processed by functions $\mathcal{R}, \mathcal{I}, \mathcal{N}$ to reshape, interpolate, and min-max normalize to the shape of the raw image ($H \times W$).

\begin{figure}[h]
\centering
 \includegraphics[width=1\textwidth]{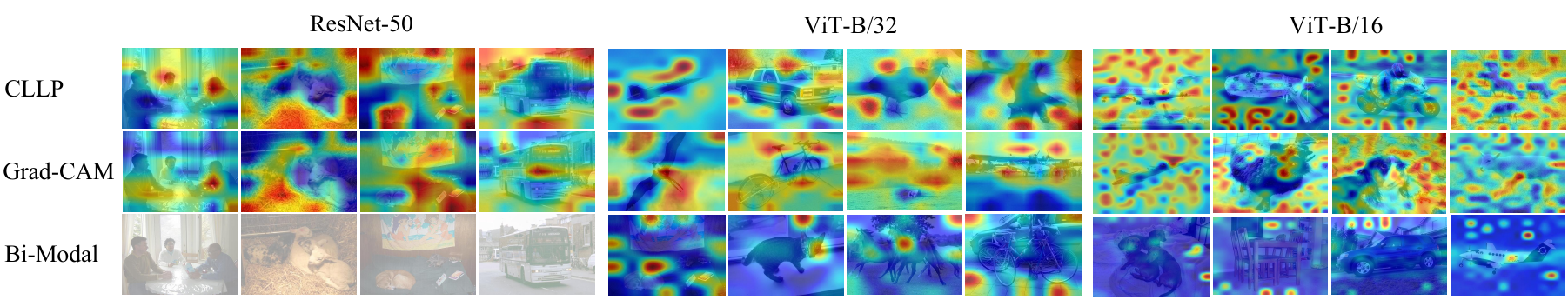}
\caption{CLIP shows opposite visualization with noisy activations, which are common on varied backbones (ResNet \cite{he2016deep}, ViT \cite{dosovitskiy2020image}) and methods (Grad-CAM \cite{selvaraju2017grad}, Bi-modal \cite{chefer2021generic}). Note, Bi-modal for ViT is not applicable to ResNet.}
\label{fig_prob_more}
\end{figure}

Subsequently, we generate similarity maps of CLIP, as shown in Figure \ref{fig_prob_more}, and observe that the most prominent problems are the opposite visualization and noisy activations. More specifically, when identifying a target category, CLIP tends to prioritize background regions over foreground regions, which contradicts human perception. Besides, there are many noisy activations at class irrelevant positions. These phenomena are observed across various backbones (ResNets \cite{he2016deep} and Vision Transformers \cite{dosovitskiy2020image}), and also occur in multiple explainability methods as shown in Fig. \ref{fig_prob_more} and Fig. \ref{fig_vis_comp}. These figures indicate these phenomena are common instead of an isolated case. 

\subsection{CLIP Architecture Surgery}\label{sec3.2}
\begin{figure}[h]
\centering
 \includegraphics[width=1\textwidth]{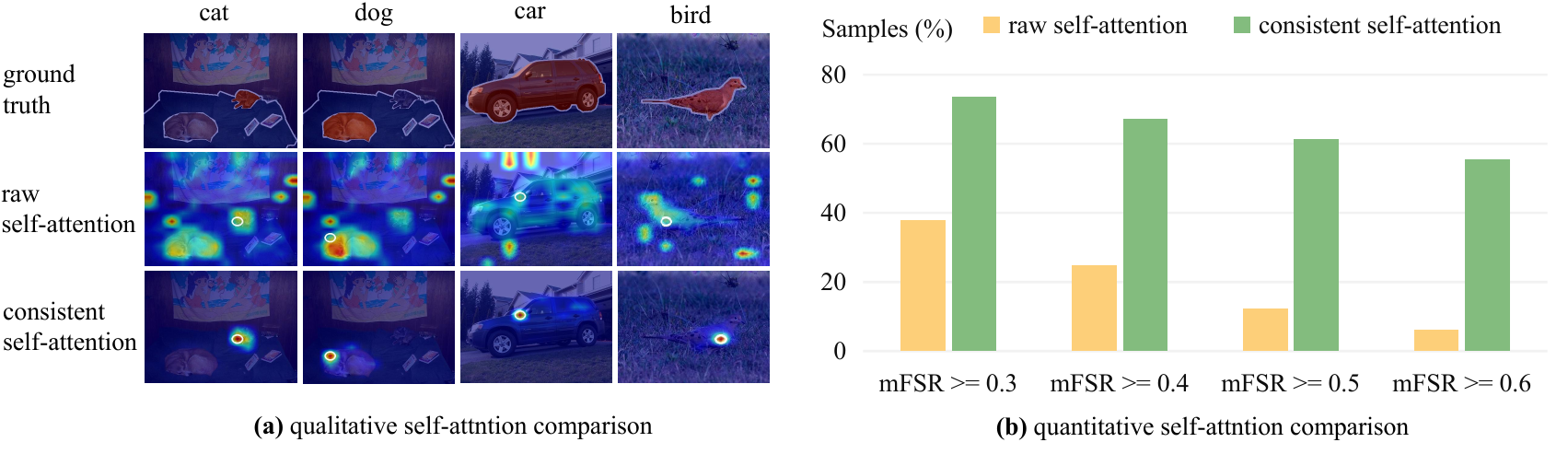}
\caption{The raw self-attention pays more attention on backgrounds, while the proposed consistent self-attention corrects it. (a) Qualitative comparison between the raw self-attention and our consistent self-attention at the token of the highest score on the last layer. The raw self-attention pays much attention to the opposite semantic regions, while our consistent self-attention links nearby tokens at the same semantics. (b) The quantitative comparison for self-attention focuses on foregrounds. Note, the y-axis indicates the sample ratio whose mFSR is higher than a threshold, and mFSR measures the attention degree on the foregrounds as Eq. \ref{eq_mFSR}.}
\label{fig_token_shift}
\end{figure}

\noindent \textbf{Consistent self-attention:} Firstly, let us draw the conclusion regarding the reason for the opposite visualization: the original self-attention layers in CLIP build relations among inconsistent semantic regions. To substantiate our claim, we present both qualitative and quantitative evidence in Fig. \ref{fig_token_shift}. In this figure, raw self-attention $\boldsymbol{A}_{raw}$ uses heterologous parameters $\phi_q,\phi_k$  which are different from the parameter $\phi_v$ for output value as:
\begin{equation}
\label{eq_attn_qk}
    \boldsymbol{A}_{raw} = \sigma(s \cdot \boldsymbol{Q} \boldsymbol{K}^{\top}) \boldsymbol{V},
\end{equation}
where $\sigma$ is the softmax function and $s$ indicates the learnable scale, $\boldsymbol{Q} = \phi_q(\boldsymbol{X})$, $\boldsymbol{K} = \phi_k(\boldsymbol{X})$, $\boldsymbol{V}=\phi_v(\boldsymbol{X})$ using learnable linear parameters $\phi_q,\phi_k,\phi_v$, respectively. 

The raw self-attention uses heterologous parameters to build global relations. However, these different parameters build relations to inconsistent semantic regions as Fig. \ref{fig_token_shift}. This is owing to the lack of local supervision to align the right relation, thus varied parameters may link to context or redundant features. After locating where the problem occurs, we proposed our solution: consistent self-attention $\boldsymbol{A}_{con}$ via homogeneous parameters as:
\begin{equation}
\label{eq_attn_vv}
    \boldsymbol{A}_{con} = \sigma(s \cdot \boldsymbol{V} \boldsymbol{V}^{\top}) \boldsymbol{V},
\end{equation}
where the self-attention matrix $\boldsymbol{V} \boldsymbol{V}^{\top}$ uses the homogeneous parameter $\phi_v$ as that for output feature.

The motivation for consistent self-attention is that the self-attention matrix based on a homogeneous parameter builds relations for consistent semantics. Specifically, there are no heterologous parameters, so the features are the same, where the token itself stands for the highest cosine similarity, and nearby or similar tokens are ranked next. Qualitative results in Fig. \ref{fig_token_shift}(a) support this claim. Besides, we introduce a metric mFSR in Eq. \ref{eq_mFSR}(b) to quantitatively prove it.

\begin{equation}
\label{eq_mFSR}
    mFSR = m_c(m_s(\frac{\sum_{i=1}^{H} \sum_{j=1}^{W} \boldsymbol{A}_{i,j} \cdot \boldsymbol{G}_{i,j}}{\sum_{i=1}^{H} \sum_{j=1}^{W} \boldsymbol{A}_{i,j}}))
\end{equation}

Specifically, mFSR quantitatively measures the ratio of attention on the foregrounds, where $m_c, m_s$ count the mean values along classes and samples (every positive label on all images), respectively. Herein, $\boldsymbol{A} \in \mathbb{R}^{H \times W} (N_t = H \times W)$ is the self-attention averaged along the head dimension from the last layer (belonging to the token at the highest score on the similarity map), and $\boldsymbol{G}$ is the foreground binary ground-truth of each sample whose size is $H\times W$. Results in Fig. \ref{fig_token_shift}(b) suggest that most examples of the proposed consistent self-attention focus on the foregrounds for varied mFSR thresholds, while a few samples are passed for the raw self-attention.

\noindent \textbf{Dual paths:} Besides the above analysis for self-attention, we observe that not all the layers are close to the final predictions and hurt the explainability. To measure the affinity $a(\cdot)$ between the final prediction and that of intermediate layers, we calculate the average cosine similarity (angle) between L2-normalized text feature $\boldsymbol{F}_t \in \mathbb{R}^{C \times N_t}$ and image feature $\hat{\boldsymbol{F}_c} \in \mathbb{R}^{C}$ at the class token of the targeted module as Eq. \ref{eq_cos_angle} (including self-attention modules and feed-forward networks (FFN) summed in the residual). Note that each image feature is multiplied with the last linear projection layer to get $\hat{\boldsymbol{F}_c}$, and the class token is used to extract image-level features for the average affinity $a(\cdot)$ among positive texts.
\begin{equation}
\label{eq_cos_angle}
    a(\boldsymbol{F}_t, \hat{\boldsymbol{F}_c}) = \frac{\sum_{i=1}^{N_t}{\boldsymbol{F}_t}^{(i)} \hat{\boldsymbol{F}_c}}{N_t}
\end{equation}

\begin{figure}[h]
\centering
 \includegraphics[width=0.75\textwidth]{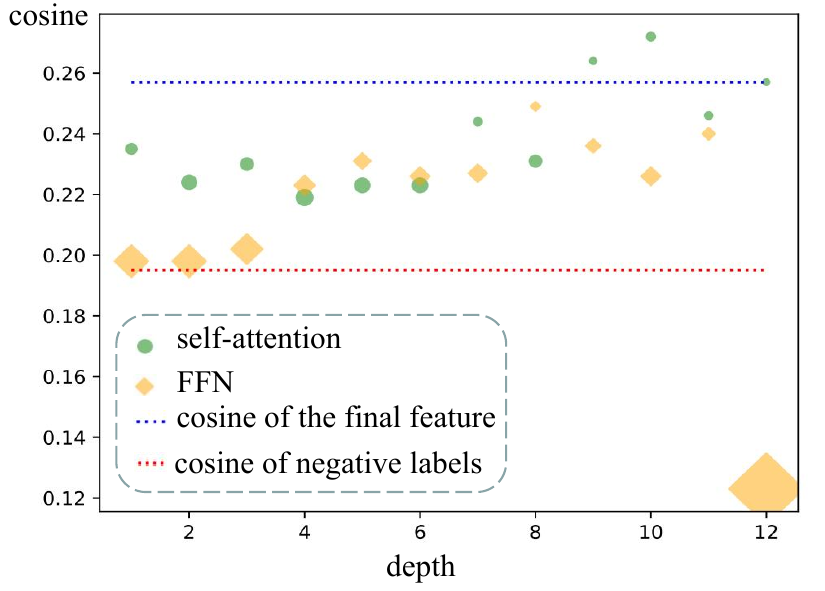}
\caption{Analysis for intermediate self-attention blocks (green) and feed-forward networks (FFNs, colored in yellow) at different depths via the affinity in Eq. \ref{eq_cos_angle}. The blue line indicates the mean cosine of all positive labels on the VOC 2012 dataset \cite{everingham2010pascal} using backbone ViT-B/16, and the red line is that of negative labels. All scatters are the mean cosine of positive labels and are expected to be close to the blue line. Larger scatters indicate the features of this block are more inconsistent with the final prediction.}
\label{fig_cos_angle}
\end{figure}

Based on this metric, we set analysis for image-level intermediate predictions for multiple blocks involved in the residual at varied depths, then we draw the affinity for each module in Fig. \ref{fig_cos_angle}. From this figure, we find FFN modules have larger gaps than self-attention modules and closed to negatives. Especially, the last FFN returns the feature at cosine 0.1231, which is much farther than the cosine of negative labels. The features of the first three FFNs are very close to the features of negative labels. This finding suggests that FFNs push features towards negatives when identifying positives, thus hurting the model. Tab. \ref{tab_dual_paths} experimentally prove this claim in the explainability task. \emph{Therefore, we only take features from partial consistent self-attention modules without FFNs.}

Guided by the above analysis, we aggregate partial self-attention modules for consistent results to the final prediction. To avoid model collapse when deleting FFN and partial layers, a technique called dual paths is proposed. Specifically, we skip FFN in the new path as analyzed in Fig. \ref{fig_cos_angle}. Then, we define the architecture surgery by the update of dual paths as Eq. \ref{eq_dual_paths} and \ref{eq_dual_paths2}. Here, $\hat{x}_{i+1}$ is the new path: 
\begin{equation}
    \label{eq_dual_paths}
    \hat{x}_{i+1} = 
    \begin{cases}
        None & i < d \\
        f_{\boldsymbol{A}_{con}}(x_i, \{\phi_v\}) + x_i & i = d \\
        f_{\boldsymbol{A}_{con}}(x_i, \{\phi_v\}) + \hat{x}_i & i > d
    \end{cases}, \ \forall T\&A. \\
\end{equation}
where $i$ is the index of the block, and depth $d$ controls the start of the new path. For shallow layers under $d$, there is only the original path, and the new path returns ``None". For the first reformed self-attention ($i=d$), we merge $x_i$ from the original path with the output of $f_{\boldsymbol{A}_{con}}(x_i, \{\phi_v\})$, which consists of the consistent self-attention in Eq. \ref{eq_attn_vv} using parameter $\phi_v$ and a linear projection layer without Feed-Forward Networks $f_{FFN}$). For the following modules $i>d$, the $x_i$ is turned to $\hat{x}_{i}$, where outputs of deeper self-attentions are merged only. This path is only applicable to Transformers and Attention Pooling ($T\&A$) of CLIP. For the original path $x_{i+1}$, the operations are not modified:
\begin{equation}
    \label{eq_dual_paths2}
    x_{i+1} = 
    \begin{cases}
    	\begin{split}
    		f&_{FFN}(x_i^{\prime}) + x_i^{\prime}, \\
    		 &\ s.t.\ \ x_i^{\prime} = f_{\boldsymbol{A}_{raw}}(x_i, \{\phi_q,\phi_k,\phi_v\}) + x_i
    	\end{split} & , \ \forall T\&A \\
    	f_{res}(x_i) + x_i & , \ \forall Res,
    \end{cases}
\end{equation}
where $f_{res}$ is residue blocks and $f_{FFN},\phi_q,\phi_k$ are kept.

\subsection{CLIP Feature Surgery} \label{sec3.3}

\begin{figure}[h]
\centering
 \includegraphics[width=0.75\textwidth]{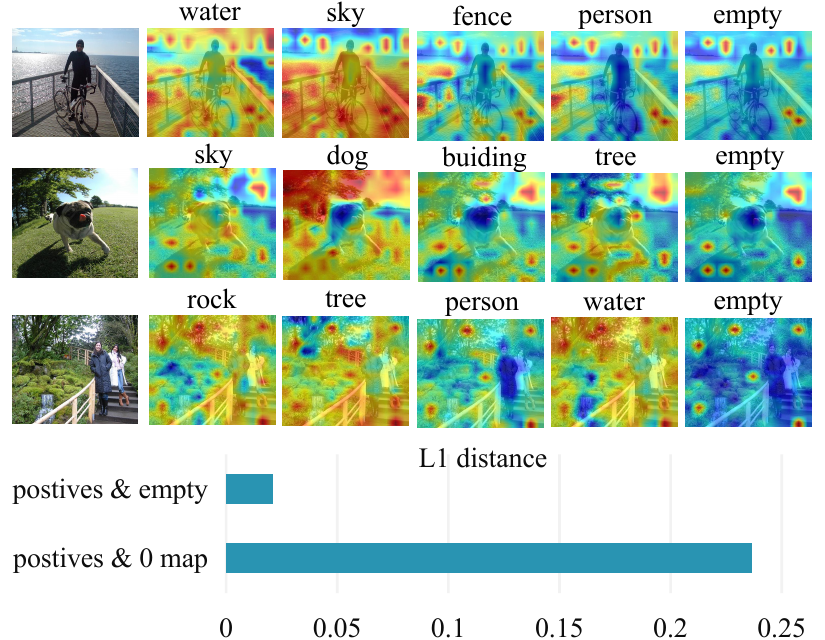}
\caption{Similarity maps from positive labels are similar to the map from an empty string, which produces redundant features irrelevant to any category. The bar chart indicates the overall L1 distance between positive maps vs. empty map or 0 map (no activation) on the PASCAL Context dataset \cite{mottaghi2014role}, suggesting the predictions of CLIP among categories are highly related to redundant features.}
\label{fig_empty_label}
\end{figure}

As shown in Fig. \ref{fig_prob_more}, the predicted similarity map of CLIP presents many noisy activations in spot shapes at unexpected positions, undermining the credibility of CLIP. We find that noisy activations are caused by redundant features among categories. Because CLIP learns to recognize numerous categories using natural language, leading to only a few features being activated for a specific class, while other features remain non-activated for the remaining classes. Consequently, these non-activated features become redundant and occupy a substantial portion of the feature space, shown as noises. The evidence is given in Fig. \ref{fig_empty_label}, where we draw the similarity maps with positive texts and an empty string. For the empty string, all the output features are regarded as redundant features without connection to any category. We can see from this figure that activations of redundant features (empty string) are very similar to noises from positive texts for both visual (top part) and quantitative results (bottom bars), which are powerful pieces of evidence.

We aim to mitigate the problem of noisy activations by removing redundant features. Motivated by the above observation, we aim to count the mean features along the class dimension to identify redundant features. Besides, in our observation, some categories are influenced by the classes at high scores, which leads to false activations. Thus, we give extra emphasis to these obvious classes, when measuring the redundant features.

In terms of formulations, we firstly obtain the multiplied features $\boldsymbol{F}_m \in \mathbb{R}^{N_i \times N_t \times C}$ in Eq. \ref{eq_fm}, by element-wise multiplication $\odot$ between the L2 normalized features $\boldsymbol{F}_i \in \mathbb{R}^{N_i \times C}$ and the normalized text features $\boldsymbol{F}_t \in \mathbb{R}^{N_t \times C}$ with expand operation $\mathcal{E}$ to broadcast in the same shape.
\begin{equation} 
\label{eq_fm}
    \boldsymbol{F}_m = \mathcal{E}(\boldsymbol{F}_{i}) \odot \mathcal{E}(\boldsymbol{F}_t)
\end{equation}
Then, we need the image similarity $\boldsymbol{s}$ of each text to pay more attention on the influential class. In Eq. \ref{eq_weight}, $\mu_{s}$ indicates the mean value of $\boldsymbol{s}$, and $s$ is obtained from the L2 normalized image feature $\boldsymbol{F}_c$ and transposed text feature $\boldsymbol{F}_t^{\top}$ with logit scale $\tau$ and softmax $\sigma$.
\begin{equation} 
\label{eq_weight}
\begin{split}
\boldsymbol{w} = &\frac{\boldsymbol{s}}{\mu_{s}} \\
    \boldsymbol{s} = &\sigma(\tau \cdot \boldsymbol{F}_{c} \boldsymbol{F}_{t}^{\top}).
\end{split}
\end{equation}
Then, we broadcast the weight $w$ and multiply to $\boldsymbol{F}_m$ in Eq. \ref{eq_fm} to count the mean value along the category dimension $N_t$ as the redundant feature $\boldsymbol{F}_r \in \mathbb{R}^{N_i \times C}$ in Eq. \ref{eq_fr}.
    \begin{equation}
        \label{eq_fr}
            \boldsymbol{F}_{r} = mean(\boldsymbol{F}_m \odot \mathcal{E}(\boldsymbol{w}))
    \end{equation}
lastly, we use multiplied features $\boldsymbol{F}_m$ to subtract the expanded redundant feature $\mathcal{E}(\boldsymbol{F}_r)$ for the removal of redundant features. Then sum all features along the channel dimension $C$ to get the cosine similarity $\boldsymbol{S} \in \mathbb{R}^{N_i \times N_t}$:
\begin{equation}
\label{eq_feature_surgery}
    \boldsymbol{S} = sum(\boldsymbol{F}_m - \mathcal{E}(\boldsymbol{F}_r))
\end{equation}

Note that Eq. \ref{eq_feature_surgery} is specifically designed for the explainability task, and the final similarity map is obtained from $\boldsymbol{M} = \mathcal{N}(\mathcal{I}(\mathcal{R}(\boldsymbol{S})))$ as Eq. \ref{eq_sm}. For multi-label recognition tasks, we only need to replace $\boldsymbol{F}_i$ in Equation \ref{eq_fm} with the image-level features of the class token $\boldsymbol{F}_c \in \mathbb{R}^{1 \times C}$ for similarity distance. Note, CLIP Feature Surgery is not suitable to rank base methods, like argmax in segmentation and top-1 accuracy in single-label classification. This is because $\boldsymbol{F}_r$ can be considered as a common bias, which does not affect the ranking of categories within a single image. Instead, it adjusts the scores across images or pixels.

%% file: 4_experiments.tex
\section{Experiments}

\subsection{Setup}

\noindent \textbf{Datasets:} The proposed CLIP Surgery is operated on the inference stage without training, thus we don't need any training datasets. To evaluate our method, we use the validation split of each dataset. Specifically, for explainability task, we use PASCAL VOC 2012 \cite{everingham2010pascal}, MS COCO 2017 \cite{lin2014microsoft}, PASCAL Context \cite{mottaghi2014role}, and ImageNet-Segmentation-50 \cite{gao2022large}, where single-label, multi-label, object and stuff are included for comprehensive evaluation. We also use the same datasets as the explainability task to evaluate the interactive segmentation task. For the open-vocabulary semantic segmentation task, besides PASCAL Context \cite{mottaghi2014role}, we test our CLIP Surgery on two widely used semantic segmentation datasets: COCO Stuff \cite{caesar2018coco} and CityScapes \cite{cordts2016cityscapes}. For the open-vocabulary multi-label recognition task, we use PASCAL Context~\cite{mottaghi2014role}, and the most used dataset NUS-Wide~\cite{chua2009nus}. Besides, we use image-text pairs in GCC3M dataset \cite{sharma2018conceptual} for the multimodal visualization task.

\begin{figure}[H]
\centering
 \includegraphics[width=1\textwidth]{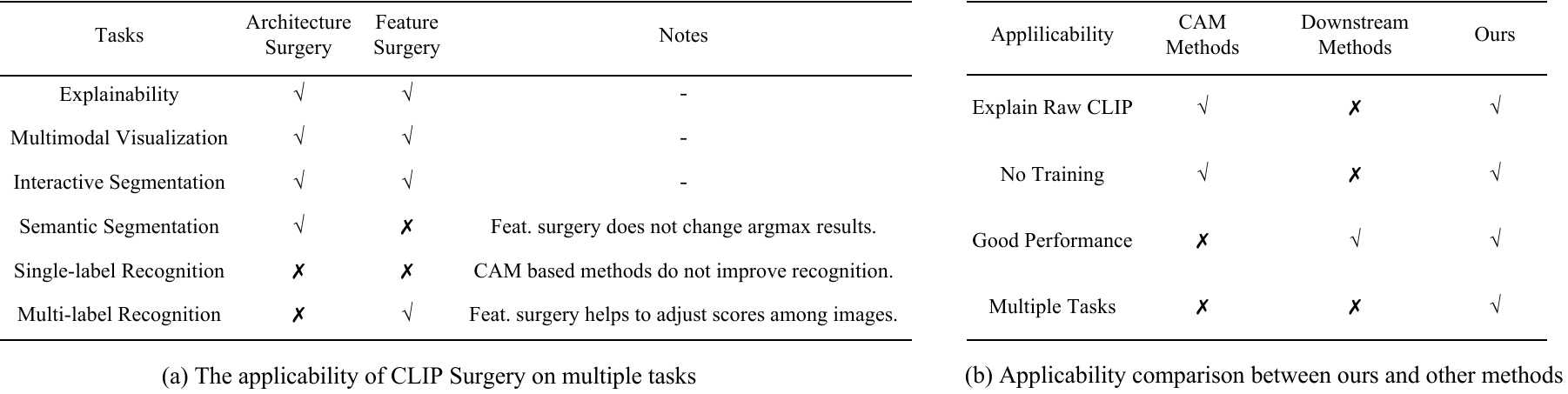}
\caption{(a) Applicability of the architecture surgery and feature surgery on varied tasks. (b) Comparison of applicability among our method, conventional CAM methods, and downstream methods.}
\label{applicability}
\end{figure}

\noindent \textbf{Implementation:} We implement the CLIP Surgery via the official CLIP. All the parameters are copied from it with modified architecture and feature operations. We elaborate the applicability of each module for varied tasks as Fig. \ref{applicability}. Specifically, semantic segmentation task and single-label recognition tasks don't apply the feature surgery, since their results are obtained from the argmax operation, while feature surgery doesn't change the prediction rank. It is also the reason why single-label recognition keeps the same results as before. Notably, it is not a shortage of our method, because \textbf{all the CAM-based methods are not responsible for classification improvements, including ours. So this paper does not involve single-label classification datasets like ImageNet}. Even though our method works in the multi-label recognition task.

For the explainability task, we set experiments on the official CLIP with 5 backbones, namely ResNet-50, ResNet-101, ViT-B/32, ViT-B/16, and ViT-L/14. For ResNets \cite{he2016deep}, CLIP adds an attention pooling with self-attention in the last layer, and their output size is 7 $\times$ 7 under input resolution 224 (all the datasets are resized to 224 $\times$ 224 without crop or other augmentation). For the output size of ViTs \cite{dosovitskiy2020image}, it depends on its patch size. Specifically, the output sizes are 7, 14, and 16 for ViT-B/32, ViT-B/16, and ViT-L/14, respectively. As for the hyperparameters of CLIP Surgery, the depth $d$ in Eq. \ref{eq_dual_paths} is set to 7 according to analysis in Fig. \ref{fig_cos_angle}, and softmax scale $\tau$ in Eq. \ref{eq_weight} is set to 2. Note that these hyperparameters are not sensitive with small variations of results. For example, on COCO 2017 \cite{lin2014microsoft} dataset using backbone ViT-B/16, the variation of results at metric mSC is 0.42\% for $d \in [1, 10]$, and variation for $\tau \in [1, 10]$ is 0.12\%. As to the textual prompt, we deploy the 85 templates in CLIP \cite{radford2021learning}, and combine templates with the names of categories. Then, we mean the text features along the template dimension as the final text features $F_t$, and this prompt ensemble is applied to all methods for fair comparison. 

For open-vocabulary semantic segmentation, we resize the images of PASCAL Context \cite{mottaghi2014role} and COCO Stuff \cite{caesar2018coco} to 512 $\times$ 512, and crop each image of Cityscapes to 8 patches at 512 $\times$ 512 from 2048 $\times$ 1024 without overlap. For fair comparisons, all the compared methods use the same backbone ViT-B/16, except grouping methods \cite{xu2022groupvit,luo2023segclip} whose backbones are specially designed. Note, the original ReCo \cite{shin2022reco} uses ResNet50x16 which is much larger than ViT-B/16, and the implemented results based on ViT-B/16 are from the official code. Since ViL-Seg \cite{liu2022open} is not released, we report its results from the paper. We reproduce the results of MaskCLIP \cite{zhou2022extract} at the same settings as ours, without its post-processing methods, for fair comparison with other works. For open-vocabulary multi-label recognition, we take the prediction from the original path (the same as the original CLIP at input size 224), and apply feature surgery to replace the softmax operation of CLIP. For other CLIP-based zero-shot methods \cite{sun2022dualcoop,xu2022dual}, we report their results from the papers. Besides, we implement TaI-DPT \cite{guo2023texts} from the official code as a baseline, and apply the feature surgery on its predictions to verify the complementarity of our method. 

For the interactive segmentation, we convert text to points for the Segment Anything Model (SAM) \cite{kirillov2023segment}. It helps to replace the cost of manual labeling and avoids the bad performance of SAM using text prompts only. Specifically, we pick points whose scores are higher than 0.8 from the similarity map, and take the same number of points ranked last as background points. Note that there is only one text prompt for SAM instead of multiple texts. Thus, we implement the feature surgery via the redundant feature $\boldsymbol{F}_{empty}$ from text features of an empty string to replace $\boldsymbol{F}_r$ in Eq. \ref{eq_feature_surgery}. This situation is the same as multimodal visualization, where the whole sentence is used as a text label without other categories. We use the string ``[start][end]" (start flag and end flag, respectively) to extract redundant features.

\noindent \textbf{Metrics:} We use mean Intersection over Union (\textbf{mIoU}) for semantic segmentation, and mean Average Precision (\textbf{mAP}) for multi-label recognition. Besides, mIoU is used to evaluate each positive label independently for interactive segmentation and explainability. The new metric for explainability is mean Score Contrast (\textbf{mSC}), which refers to the score difference between foregrounds and backgrounds, ranging from -100\% to 100\%. This metric can reflect the problem of opposite visualization when the value is lower than 0, but mIoU cannot. Besides, mSC measures the difference of scores, while mIoU measures the mask without information of confidence, which provides new insights. We give the formula definition as:
\begin{equation}
\label{eq_msc}
mSC = m_c(m_s(m_p(\boldsymbol{M}^s \cdot \boldsymbol{G}) - m_p(\boldsymbol{M}^s \cdot \neg\boldsymbol{G}))) \ ,
\end{equation}
where $m_c, m_s, m_p$ compute the mean values along classes (macro average), samples (the average over all images for a class), and pixels (spatial average), respectively. Note, $\boldsymbol{M}^s$ is the similarity map for a certain sample, $\boldsymbol{G}$ is the corresponding binary foreground ground-truth, and $\neg\boldsymbol{G}$ indicates the background ground-truth.

\subsection{Results of Explainability}

\begin{table}[H]
\centering
\caption{\label{tab_comp_clip}Results comparison between ``CLIP" and ``Ours" on four datasets and five backbones. mSC (\%) ranges from -100\% to 100\%, reflecting the score contrast between foreground and background. Note, ``$\Delta$" indicates our average improvements over five backbones.}
\setlength\tabcolsep{4.5pt}
\begin{tabular}{cccccccccc}
\hline 
  &  & \multicolumn{2}{c}{ImageNet-S50} &\multicolumn{2}{c}{VOC 2012} & \multicolumn{2}{c}{PASCAL Context} & \multicolumn{2}{c}{COCO 2017}\\
Method & Network & mIoU $\uparrow$ & mSC $\uparrow$ & mIoU $\uparrow$ & mSC $\uparrow$ & mIoU $\uparrow$ & mSC $\uparrow$ & mIoU $\uparrow$ & mSC $\uparrow$ \\
\hline
CLIP & ResNet50 & 28.18 & -26.50 & 17.78 & -27.88 & 16.98 & -14.02 & 10.53 & -19.17\\
\textbf{Ours} & ResNet50 & 66.05 & 43.28 & 53.85 & 44.60 & 38.50 & 32.44 & 29.24 & 33.80 \\
\cdashline{1-10}[0.5pt/5pt]
CLIP & ResNet101 & 28.17 & -23.30 & 18.06 & -23.48 & 17.52 & -9.90 & 10.66 & -17.31 \\
\textbf{Ours} & ResNet101 & 65.51 & 43.22 & 52.51 & 44.07 & 38.03 & 32.21 & 29.89 & 35.50\\
\cdashline{1-10}[0.5pt/5pt]
CLIP & ViT-B/32 & 28.05 & -21.86 & 17.56 & -24.06 & 16.37 & -17.68 & 10.15 & -21.02\\
\textbf{Ours} & ViT-B/32 & 59.24 & 36.72 & 51.14 & 41.15 & 40.10 & 32.90 & 29.22 & 31.35\\
\cdashline{1-10}[0.5pt/5pt]
CLIP & ViT-B/16 & 27.87 & -18.84 & 17.36 & -19.86 & 15.76 & -16.73 & 9.74 & -23.37\\
\textbf{Ours} & ViT-B/16 & 62.41 & 36.50 & 55.78 & 41.64 & 46.28 & 34.32 & 35.23 & 35.43\\
\cdashline{1-10}[0.5pt/5pt]
CLIP & ViT-L/14 & 27.89 & -18.34 & 17.24 & -24.42 & 15.62 & -20.26 & 9.64 & -27.51 \\
\textbf{Ours} & ViT-L/14 & 61.72 & 28.25 & 54.47 & 34.89 & 42.71 & 28.11 & 37.67 & 32.54\\
\hline 
$\Delta$ & - & \textbf{34.96} & \textbf{59.36} & \textbf{35.95} & \textbf{65.21} & \textbf{24.67} & \textbf{47.72} & \textbf{22.11} & \textbf{55.40} \\
\hline 
\end{tabular}
\end{table}

\noindent \textbf{Effectiveness:} Table \ref{tab_comp_clip} shows the comparison between the proposed CLIP Surgery and the original CLIP model. The comparison is performed on four datasets using five backbones. Our results consistently outperform the CLIP baselines in each setting, with significant improvements. On average, our method achieves a higher mIoU than CLIP by 22.11\% to 35.95\%, and a higher mSC by 47.72\% to 65.21\%. Notably, the mSC of CLIP is lower than 0, indicating that the model tends to favor the background over the foreground, our CLIP Surgery corrects this issue, yielding results far above 0. These results provide strong evidence for the effectiveness of our proposed method.

\begin{figure}[H]
\centering
 \includegraphics[width=1\textwidth]{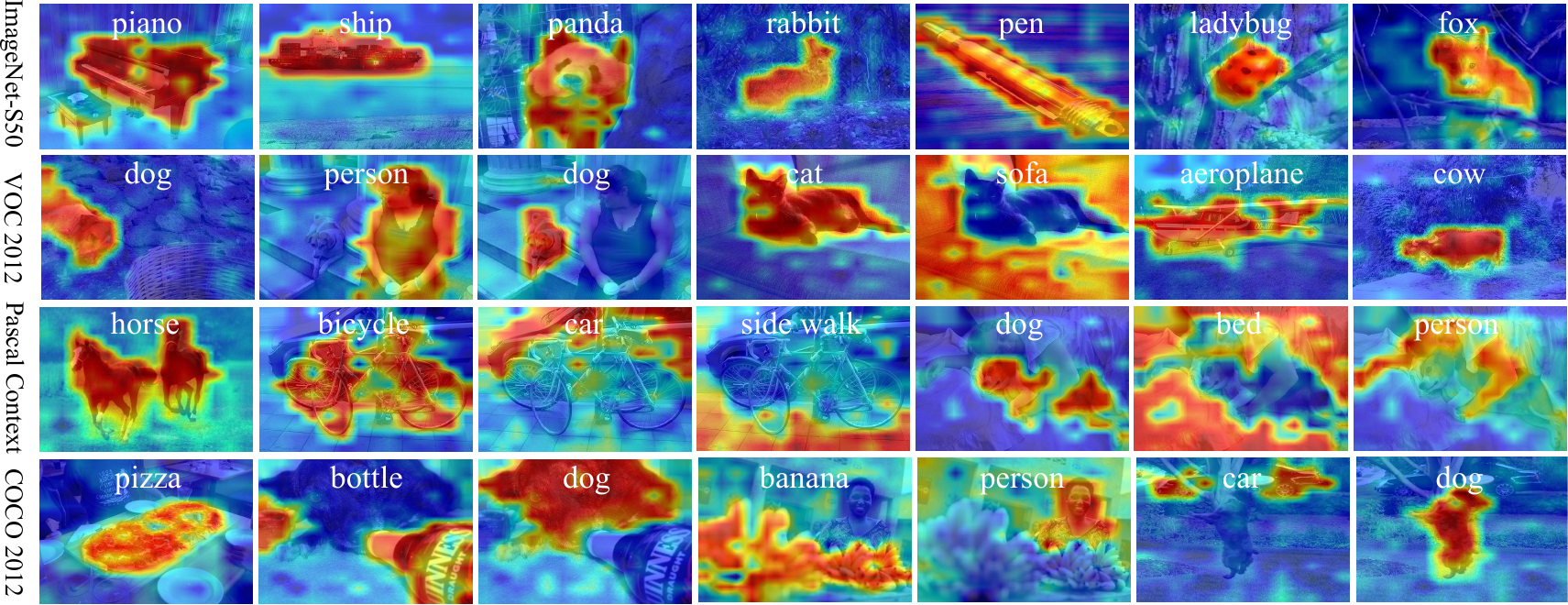}
\caption{Our CLIP Surgery solves the explainability problems and provides good visualizations on varied datasets.}
\label{fig_vis_datasets}
\end{figure}

Besides quantitative results, we draw visualization results for different datasets in Fig. \ref{fig_vis_datasets}. These results demonstrate that our proposed method effectively addresses the two discussed problems: opposite visualization and noisy activations. Notably, the proposed method enables us to produce clear and interpretable visualizations based solely on the original CLIP, without any additional training or complex back-propagation.

\noindent \textbf{Compared with Previous Works:} We compare CLIP Surgery with previous explainability methods, including the similarity map of original CLIP, Grad-CAM \cite{selvaraju2017grad} for CNN, pLRP \cite{lapuschkin2019unmasking} implemented by Chefer et al. \cite{chefer2021transformer}, Bi-Modal based on ViT, and explainability methods by \cite{li2022exploring,chen2022gscorecam}. We conduct the comparison using the ViT-B/16 and ResNet-50 from the official codebase. Tab. \ref{tab_comp_sota} displays the results, where CLIP Surgery achieves the best performance across all datasets, metrics, and backbones, surpassing other methods by significant margins. ECLIP \cite{li2022exploring} ranks second for ViT-B/16, but our CLIP Surgery outperforms it by a maximum of 15.94\% in mIoU and 19.89\% in mSC without any training. RCLIP \cite{li2022exploring} is the top method without extra training, and our results surpass it by 18.42\% in mIoU and 23.13\% in mSC. Other methods exhibit larger performance gaps compared to ours. Notably, some methods \cite{lapuschkin2019unmasking,chefer2021generic} for ViT are not applicable to ResNet, while we also achieve much higher results on ResNet-50 in this table.

\begin{table}[H]
\caption{\label{tab_comp_sota} Results compared with previous state-of-the-art explainability methods. Besides, ``CLIP'' indicates the similarity map of CLIP. ```-'' means this method is not applicable to ResNet, ``\dag" notes the model requires extra fine-tuning, and ``$\star$" indicates that back-propagations are required for each label at lower efficiency. Note, mIoU (\%) measures the mean intersection over union for positive labels, and mSC indicates the mean score contrast in Eq. \ref{eq_msc}.}
\setlength\tabcolsep{4pt}
\centering
\begin{tabular}{ccccccccc}
\hline 
  &  \multicolumn{2}{c}{ImageNet-S50} &\multicolumn{2}{c}{VOC 2012} & \multicolumn{2}{c}{PASCAL Context} & \multicolumn{2}{c}{COCO 2017}\\
Method & mIoU $\uparrow$ & mSC $\uparrow$ & mIoU $\uparrow$ & mSC $\uparrow$ & mIoU $\uparrow$ & mSC $\uparrow$ & mIoU $\uparrow$ & mSC $\uparrow$ \\
\hline
\multicolumn{9}{c}{ResNet-50} \\
\cdashline{1-9}[0.5pt/5pt]
CLIP \cite{radford2021learning} & 28.18 & -26.50 & 17.78 & -27.88 & 16.98 & -14.02 & 10.53 & -19.17\\
Grad-CAM$^\star$ \cite{selvaraju2017grad} & 34.39 & 1.30 & 22.93 & 1.76 & 19.85 & 0.76 & 13.11 & 1.89 \\
pLRP$^\star$ \cite{lapuschkin2019unmasking} & - & - & - & - & - & - & - & - \\
Bi-Modal$^\star$ \cite{chefer2021generic} & - & - & - & - & - & - & - & - \\
gScoreCAM$^\star$ \cite{chen2022gscorecam} & 62.21 & 33.80 & 48.53 & 33.69 & 34.68 & 23.27 & 12.98 & 13.32 \\
RCLIP \cite{li2022exploring} & 54.45 & 26.50 & 41.17 & 27.88 & 28.48 & 14.02 & 21.87 & 19.17 \\
ECLIP\dag \cite{li2022exploring} & 62.49 & 30.49 & 50.04 & 31.14 & 36.28 & 21.77 & 27.27 & 23.43 \\
\textbf{CLIP Surgery (Ours)} & \textbf{66.05} & \textbf{43.28} & \textbf{53.85} & \textbf{44.60} & \textbf{38.50} & \textbf{32.44} & \textbf{29.24} & \textbf{33.80} \\
\hline
\multicolumn{9}{c}{ViT-B/16} \\
\cdashline{1-9}[0.5pt/5pt]
CLIP \cite{radford2021learning} & 27.87 & -18.84 & 17.36 & -19.86 & 15.76 & -16.73 & 9.74 & -23.37\\
Grad-CAM$^\star$ \cite{selvaraju2017grad} & 28.59 & -11.05 & 17.90 & -14.51 & 16.04 & -14.68 & 9.89 & -18.91 \\
pLRP$^\star$ \cite{lapuschkin2019unmasking} & 46.76 & 10.33 & 31.73 & 8.88 & 25.61 & 6.24 & 21.06 & 11.22 \\
Bi-Modal$^\star$ \cite{chefer2021generic} & 43.37 & 6.77 & 30.64 & 6.76 & 24.31 & 3.95 & 18.33 & 7.99 \\
gScoreCAM$^\star$ \cite{chen2022gscorecam} & 24.75 & 7.47 & 11.33 & 1.59 & 13.26 & 0.35 & 13.59 & 4.45 \\
RCLIP \cite{li2022exploring} & 48.00 & 16.14 & 37.36 & 18.51 & 33.25 & 18.21 & 26.12 & 22.41 \\
ECLIP\dag \cite{li2022exploring} & 58.59 & 26.32 & 48.46 & 28.83 & 30.34 & 14.43 & 24.67 & 18.95 \\
\textbf{CLIP Surgery (Ours)} & \textbf{62.41} & \textbf{36.50} & \textbf{55.78} & \textbf{41.64} & \textbf{46.28} & \textbf{34.32} & \textbf{35.23} & \textbf{35.43} \\
\hline 
\end{tabular}
\end{table}

\begin{figure}[H]
\centering
 \includegraphics[width=1\textwidth]{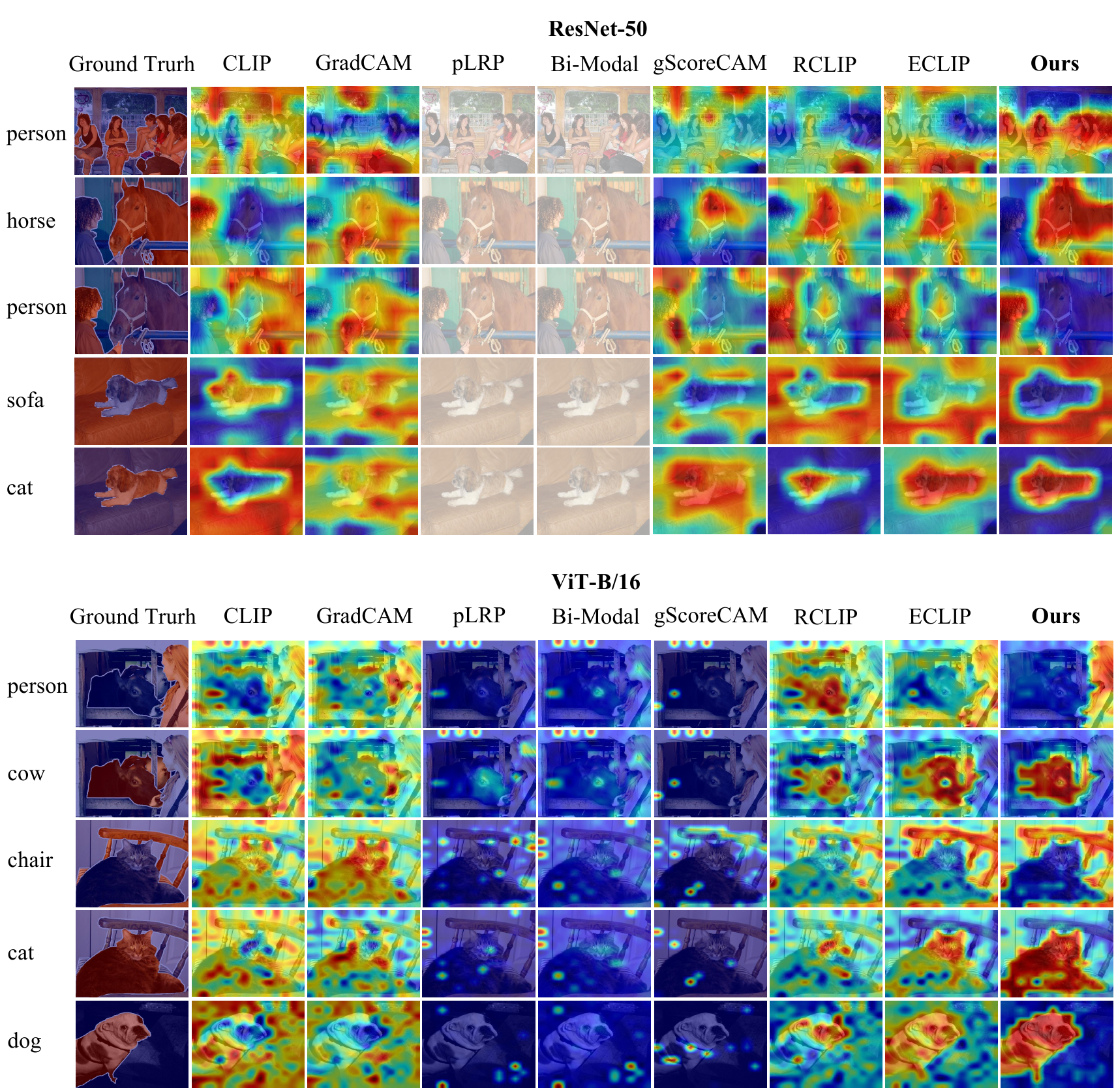}
\caption{Visual comparison between our CLIP Surgery with state-of-the-art explainability methods on VOC 2012 \cite{everingham2010pascal}. Note the foregrounds are colored in red, and the white mask indicates this method is not applicable to this backbone. Our visualization quality is much better than other methods for both ResNet and ViT, without any fine-tuning like ECLIP \cite{li2022exploring} or back-propagation in GradCAM \cite{selvaraju2017grad}. Besides, our method is not limited by certain backbones such as pLRP \cite{lapuschkin2019unmasking}, Bi-Modal \cite{chefer2021generic}, and gScoreCAM \cite{chen2022gscorecam}.}
\label{fig_vis_comp}
\end{figure}

In addition, we also present a visual comparison in Fig. \ref{fig_vis_comp}. The results show that our CLIP Surgery provides better quality visualizations compared to existing explainability methods, without problems of opposite visualization. Furthermore, our method produces fewer noisy activations than methods like pLRP \cite{lapuschkin2019unmasking}, Bi-Modal \cite{chefer2021generic}, gScoreCAM \cite{chen2022gscorecam}. Also, our method shows a more obvious score contrast than RCLIP and ECLIP \cite{li2022exploring} at better visualization quality. All the above improvements enhance the model's visual explainability and make it more credible.

\subsection{Ablation Study}

\begin{table}[H]
\centering
\caption{\label{tab_feature_surgery} Ablation study of CLIP Surgery. ``no" is the original CLIP, ``Archi." indicates the CLIP Architecture Surgery and ``Feat." means the CLIP Feature Surgery. Note, mSC lower than 0 suggests scores of background is higher than foreground. The used dataset is the Pascal Context dataset \cite{mottaghi2014role} using CLIP with a ViT-B/16 backbone.}
\setlength\tabcolsep{20pt}
\begin{tabular}{c|cc|c}
\hline 
& \multicolumn{2}{c|}{Explainability} & Multi-label \\
CLIP Surgery & mSC (\%) $\uparrow$ & mIoU (\%) $\uparrow$ & mAP (\%) $\uparrow$ \\
\hline
no & -16.73 & 15.76 & 47.09 \\
Archi.  & 31.15 & 43.47 & 47.09 \\
Archi. + Feat. & 34.32 & 46.28 & 52.61 \\
\hline 
\end{tabular}
\end{table}

We conduct ablation studies on the Pascal Context dataset \cite{mottaghi2014role} using CLIP with a ViT-B/16 backbone. Tab. \ref{tab_feature_surgery} presents the quantitative results of the ablation experiment for two aspects of CLIP Surgery. CLIP Architecture Surgery enhances explainability by 47.88\% based on the mSC metric, while CLIP Feature Surgery further improves it by 3.17\%. For the multi-label recognition task with features from the original path, the mAP remains unchanged. However, CLIP Feature Surgery significantly enhances it by 5.52\%. These results indicate that feature surgery improves both the explainability and multi-label recognition tasks.

Tab. \ref{tab_dual_paths} presents the ablation study on multiple layers, dual paths, and FFN. Replacing the last raw self-attention (Eq. \ref{eq_attn_qk}) with the new self-attention (Eq. \ref{eq_attn_vv}) results in an mSC of 29.13\% (row 1). However, applying the new self-attention to multiple layers leads to a decline of -7.73\% due to the modified outputs causing model instability. This issue is resolved by the proposed dual paths, which stabilize the model at an mSC of 32.30\% (40.03\% higher than the original single path). The FFN alone is ineffective without self-attention, while the new self-attention performs better without FFN (-4.28\% vs. 34.32\%). The results highlight the importance of dual paths for multi-layers, and using the new self-attention alone achieves the best result at 34.32\%.

\begin{table}[H]
\centering
\caption{\label{tab_dual_paths} Ablation study about multi-layers, dual paths, and FFN. ``Last" indicates the Archi. Surgery is only applied on the last layer, and ``Multi" means applying to multiple layers. ``Dual" is the proposed dual paths. Besides, ``FFN" indicates only feed-forward networks are used, and ``Attn" only takes self-attentions.}
\setlength\tabcolsep{18pt}
\begin{tabular}{ccccc|c}
\hline 
Last & Multi & Dual & FFN & Attn & mSC (\%) $\uparrow$ \\
\hline
\usym{2613} & \usym{2613} & \usym{2613} & \usym{2613} & \usym{2613} & -16.73 \\
\checkmark & \usym{2613} & \usym{2613} & \usym{2613} & \usym{2613} & 29.13 \\
\usym{2613} & \checkmark & \usym{2613} & \usym{2613} & \usym{2613} & -7.73 \\
\usym{2613} & \checkmark & \checkmark & \usym{2613} & \usym{2613} & 32.30 \\
\usym{2613} & \checkmark & \checkmark & \checkmark & \usym{2613} & -4.28 \\
\usym{2613} & \checkmark & \checkmark & \usym{2613} & \checkmark & 34.32 \\
\hline 
\end{tabular}
\end{table}

\subsection{Results on Open-vocabulary Tasks}

\begin{table}[H]
\footnotesize
\centering
\caption{\label{tab_seg}Our method helps the raw CLIP archives comparison results with some SoTA open-vocabulary semantic segmentation without any further training or alignment. Besides, we list more differences in Fig. \ref{applicability}(b) to compare these task-specific downstream methods. All methods use the same backbone ViT-B/16, without box supervision, post-processing, or self-training for fair comparison at metric mIoU (\%).}
\setlength\tabcolsep{2pt}
\begin{tabular}{ccccccc}
\hline 
Method & Weights & Training & Core Idea & PASCAL Context & COCO Stuff & Cityscapes \\
\hline 
ViL-Seg \cite{liu2022open} & CLIP & yes & contrasting and clustering & 16.3 & 16.4 & - \\
OVSegmentor \cite{xu2023learning} & CLIP & yes & learnable group tokens & 20.4 & - & - \\
ReCo \cite{shin2022reco} & CLIP & yes & co-segmentation with retrieval & 22.3 & 14.8 & 21.1\\
GroupViT \cite{xu2022groupvit} & scratch & yes & tokens grouping & 22.4 & 13.3 & 12.4 \\
Seg-CLIP \cite{luo2023segclip} & CLIP & yes & grouping with learnable centers & 24.7 & - & - \\
ZeroSeg \cite{Chen_2023_ICCV} & CLIP & yes & visual distillation & 20.4 & 20.2 & - \\
MaskCLIP \cite{zhou2022extract} & CLIP & no & modify attention pooling & 22.4 & 15.3 & 22.7 \\
\textbf{CLIP Surgery} & CLIP & no & improve explainability & \textbf{29.3} & \textbf{21.9} & \textbf{31.4} \\
\hline 
\end{tabular}
\end{table}

\noindent \textbf{Semantic Segmentation:} We find that the high-quality visualization results from the CLIP Surgery method are well-suited for the task of semantic segmentation. Despite requiring no extra training, our method performs exceptionally well, and even surpasses some open-vocabulary segmentation methods that require extra training. As shown in Tab. \ref{tab_seg}, our CLIP Surgery achieves the best results on three datasets, surpassing the second-best method by 4.6\%, 1.7\%, and 8.7\%, respectively. Besides the noticeable effectiveness, we list the differences of core ideas among methods, and CLIP Surgery is the prior work to introduce explainability into open-vocabulary segmentation task with high novelty. Note, there are many other open-vocabulary segmentation methods listed in the related work. While they require partial mask annotations \cite{xu2021simple}, additional supervised proposal network \cite{zhong2022regionclip}, extra supervision \cite{zhang2023simple}, etc. Thus, we do not compare them in Tab. \ref{tab_seg}.

\noindent \textbf{Multi-label Recognition:} We compare our method with existing CLIP-based zero-shot multi-label recognition in Tab. \ref{tab_comp_multi}. Our method significantly improves the mAP of the raw CLIP by 11.61\% and 7.24\% for ViT-B/16 and ResNet-50, respectively, on the NUS-Wide \cite{chua2009nus} dataset. Besides, our results have already beyond some methods requiring fine-tuning. After combining with TaI-DPT \cite{guo2023texts}, we achieve new state-of-the-art result on NUS-Wide \cite{chua2009nus} at mAP 48.55\% using ResNet-50, which is 1.56\% higher than the implemented TaI-DPT from its official code. These results suggest our method is effective, also it's complementary to methods at zero-shot settings which require fine-tuning on seen categories.

\begin{table}[H]
\centering
\caption{\label{tab_comp_multi} Results of CLIP-based multi-label recognition on NUS-Wide dataset \cite{chua2009nus}. Our method is complementary to zero-shot methods in the bottom part.}
\setlength\tabcolsep{18pt}
\begin{tabular}{cccc}
\hline 
Method & Network & Fine-tuning & mAP (\%) \\
\hline
CLIP \cite{radford2021learning} & ViT-B/16 & no & 35.58 \\
Ours & ViT-B/16 & no & \textbf{47.19} \\
\cdashline{1-4}[0.5pt/5pt]
CLIP \cite{radford2021learning} & ResNet-50 & no & 32.75 \\
\textbf{Ours} & ResNet-50 & no & \textbf{39.99} \\
\hline
DualModal \cite{xu2022dual} & ViT-B/32 & yes & 36.56 \\
MKT \cite{he2023open} & ViT-B/16 & yes & 37.6 \\
DualCoOp \cite{sun2022dualcoop} & ResNet-50 & yes & 43.6 \\
TaI-DPT \cite{guo2023texts} & ResNet-50 & yes & 46.99 \\
\textbf{TaI-DPT + Ours} & ResNet-50 & yes & \textbf{48.55} \\
\hline 
\end{tabular}
\end{table}

\begin{figure}[H]
\centering
 \includegraphics[width=1\textwidth]{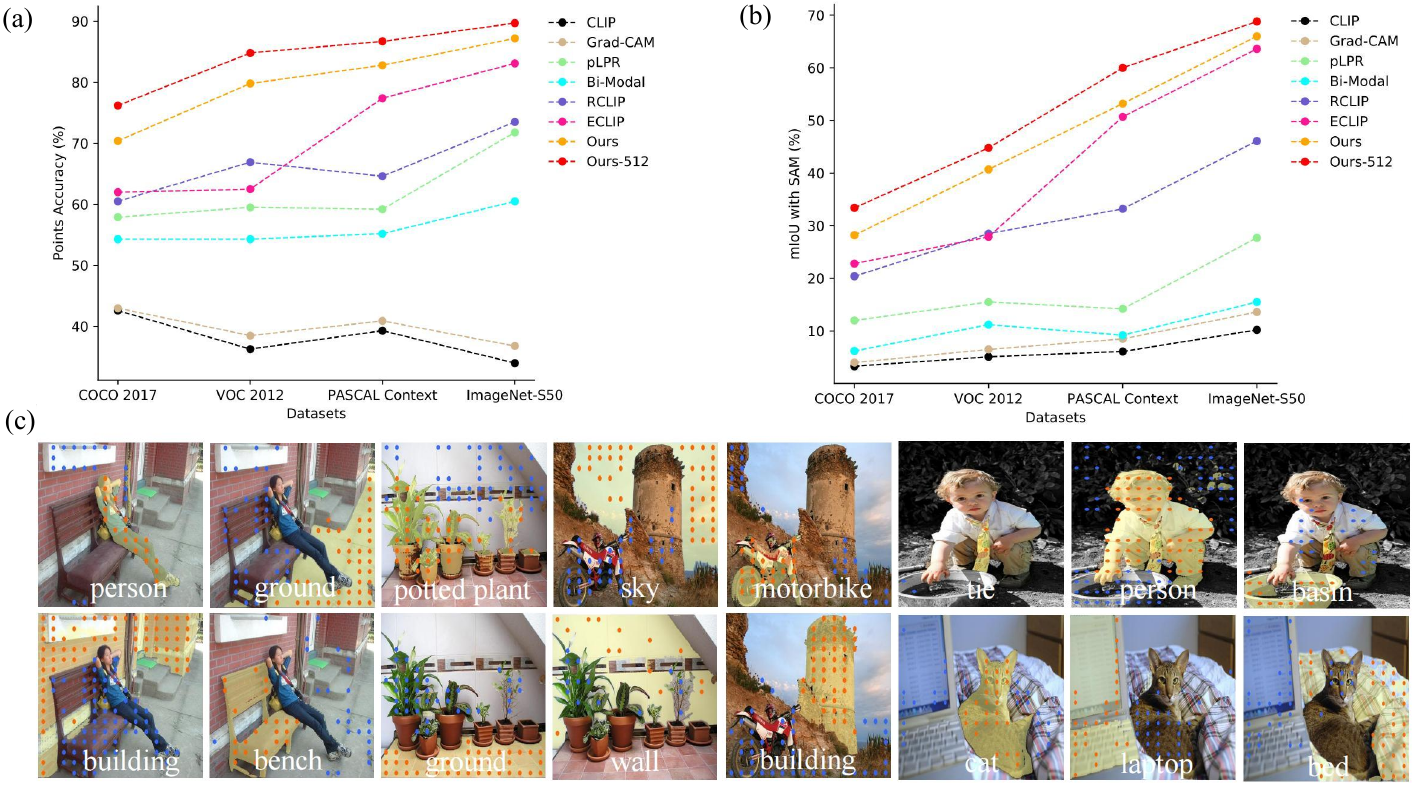}
\caption{Comparison of points accuracy (a) and mIoU (b) with SAM \cite{kirillov2023segment} among explainability methods on four datasets. Points and masks of our method are shown in (c). Note, all the input size is 224, except ``Ours-512" at 512.}
\label{fig_sam}
\end{figure}

\noindent \textbf{Interactive Segmentation:} Interactive segmentation \cite{ngc2023dahu,kirillov2023segment} involves segmenting a target object from an image with user guidance. \textbf{Segment Anything Model (SAM)} \cite{kirillov2023segment} is a new paradigm of it. SAM enables interactive segmentation via text prompts in an open-vocabulary manner. However, it performs poorly with text prompts alone, and the authors suggest combining text with manual points for better results. Our motivation is to replace the need for manual points entirely by using CLIP Surgery with text-only inputs. Our proposed method provides pixel-level results from text input (orange points in Fig. \ref{fig_sam} indicate predicted foregrounds and blue points are background prompts), which can be readily converted to point prompts for the SAM model. The advantages include (1) low manual efforts without combination with manual point; (2) one text for all images without interactions on every image;

We compare our method with other explainability methods in the \textbf{text-to-points solution}. Using ViT-B/16 as the backbone, we select points with scores above 0.8 as positives and use the same number of lowest-ranked negatives as input prompts for SAM. Fig. \ref{fig_sam}(b) illustrates the mIoU after SAM processing and the qualitative results, and Fig. \ref{fig_sam}(a) shows the accuracy of points compared with other CAM-based methods. Note that the mIoU is evaluated independently for each positive label. Our method outperforms others at large margins and performs well visually. Especially, there are over three times improvements compared with the original CLIP and explainability methods for CNN and ViT.

\noindent \textbf{Multimodal Visualization:} 
Besides above visualization results on image modality for varied tasks, we also explain the learning process of CLIP by the multimodal visualization. Specifically, we visualize the image-text pairs during training, where the whole sentence is used as a textual label for visual explanation. At the same time, we show the top text tokens whose scores are ranked in front. For the implementation, we use image-text pairs (training data) from the GCC3M dataset \cite{sharma2018conceptual}, since the training data of CLIP is private and not available. Then we draw the similarity map via our CLIP Surgery method and mark the high-response text tokens. Specifically, the feature of the class token $\boldsymbol{F_c}$ is used to compute similarity scores for each text token, and the text at max similarity is served for the generation of similarity map with image tokens $\boldsymbol{F_i}$.

\begin{figure}[H]
\centering
 \includegraphics[width=0.9\textwidth]{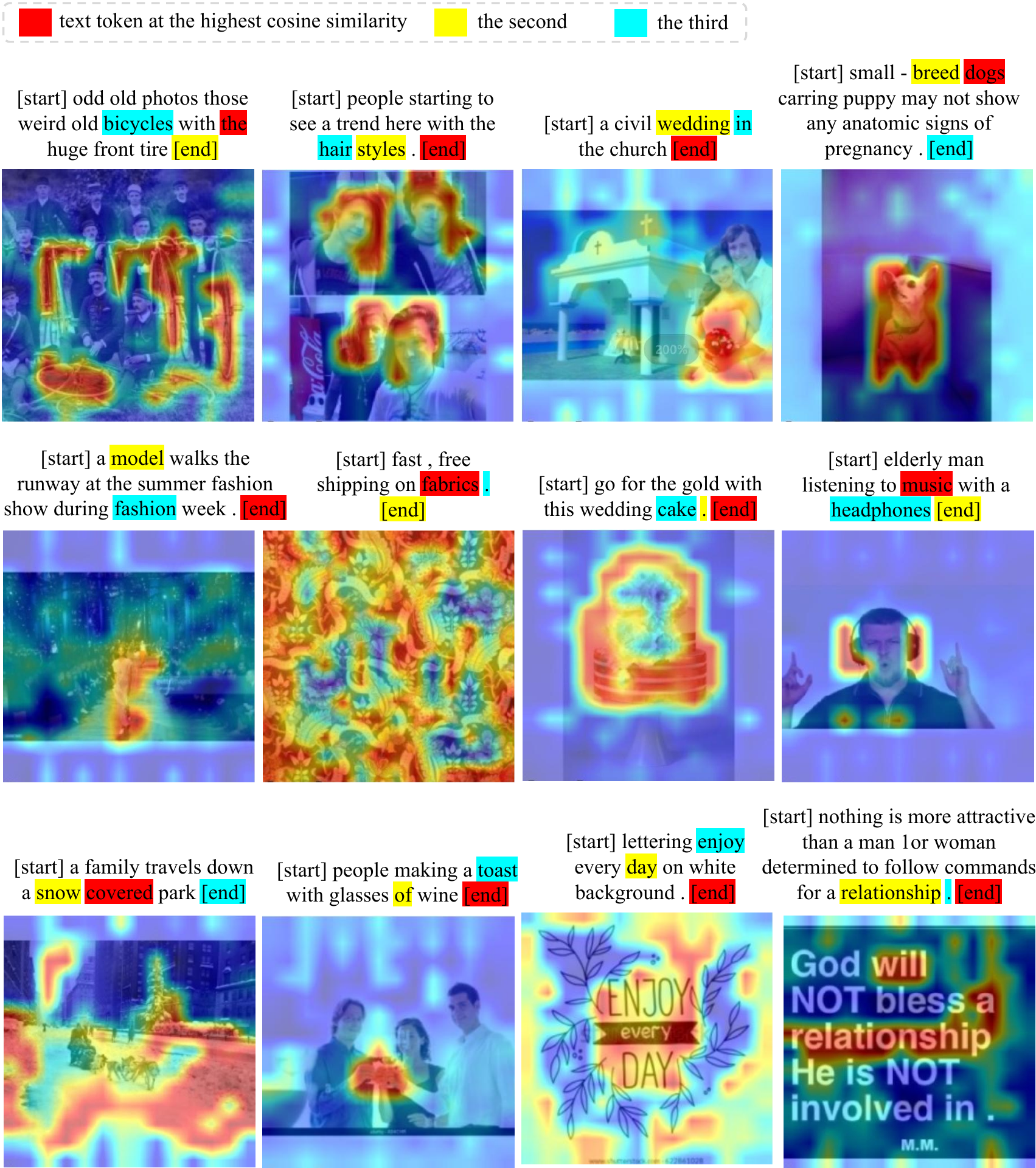}
\caption{Multimodal visualization to explain the image-text pairs used in the training process of CLIP. The visual explainability results are from the proposed CLIP Surgery, and textual explainability results are based on the cosine distance of each text token. [start] indicates the start text token and [end] means the end text token. Note that we mark the text token at the highest cosine similarly in red, then draw the second in yellow and blue for the third.}
\label{fig_vis_multimodal}
\end{figure}

We draw multimodal visualization results as Fig. \ref{fig_vis_multimodal}. From these multimodal visualization results, we observed some interesting phenomena. For the visual results, we summarize two points: (1) Not all the objects or stuff are highlighted in the image, because only the text token at the highest cosine similarity is picked for training (red texts in Fig. \ref{fig_vis_multimodal}). Thus, we believe CLIP learns partial context from one image. (2) CLIP can recognize the texts from an image to some extent, as shown in the last two images of Fig. \ref{fig_vis_multimodal}. Since the highlights are corresponded with text tokens (e.g., day, enjoy, relationship). For the textual visualization results, there are two findings: (1) The end token is the most common activated text token, and some non-object words are at high response too (e.g., ``in", ``.", ``of"). It suggests there are also redundant tokens in the vocabulary dictionary. (2) The object-based words also occur frequently with corresponding salient objects in the image. While their cosine similarities often rank second or third behind the end token ``[end]". These findings are interesting, also they reveal some characteristics of image-text pairs of CLIP, thus providing potential value to further improvement of CLIP's training process.

%% file: 5_conclusion.tex
\section{Conclusion}
In this study, we investigate two observations related to CLIP's explainability: opposite visualization and noisy activations. We discover that the raw self-attentions build relations on inconsistent semantic regions, resulting in the opposite visualization. To address it, we propose the CLIP architecture surgery, merging the consistent self-attentions with a dual paths structure. For the noisy activation in CLIP, it arises from redundant features among categories, then we introduce the CLIP feature surgery on output features to mitigate the common but redundant activations.

The proposed method significantly enhances the visual explainability of CLIP for reliable CAM, which plays a crucial role in promoting model transparency. Moreover, our method further enhances downstream tasks like semantic segmentation, interactive segmentation, and multi-label recognition, showing remarkable improvements. Besides, it provides valuable insights into the architecture, features, and learning process of CLIP, which boosts our understanding of CLIP and benefits it further improvements. Overall, the proposed CLIP Surgery offers a promising solution to generate high-quality CAM for CLIP, with wide applicability and valuable insights.
\\

\noindent \textbf{CRediT authorship contribution statement}

\textbf{Yi Li}: Conceptualization, Investigation, Methodology, Validation, Visualization, Writing. \textbf{Hualiang Wang}: Conceptualization, Investigation, Methodology. \textbf{Yiqun Duan}: Conceptualization, Investigation. \textbf{Jiheng Zhang}: Resources. \textbf{Xiaomeng Li}: Conceptualization, Investigation, Project administration, Supervision, Writing – review \& editing.

\noindent \textbf{Declaration of competing interest}

We declare that this manuscript has not been published before and is not currently being considered for publication elsewhere. We confirm that the manuscript has been approved by all authors for publication, and no conflict of interest exits in the submission of it.

\noindent \textbf{Data availability}

All data used in the research are publicly available.

\noindent \textbf{Acknowledgments}

This work is partially supported by the National Natural Science Foundation of China under Grant 62306254, and also by grants from Foshan HKUST Projects under Grants FSUST21-HKUST10E and FSUST21-HKUST11E.